\documentclass[conference]{IEEEtran}
\IEEEoverridecommandlockouts
\usepackage{cite}
\usepackage{amsmath,amssymb,amsfonts}
\usepackage{algorithmic}
\usepackage{graphicx}
\usepackage{textcomp}
\usepackage{xcolor}
\usepackage{subcaption}
\usepackage{multirow}
\usepackage{booktabs}
\usepackage{todonotes}

\usepackage[colorlinks,
linkcolor=blue,
anchorcolor=blue,
citecolor=blue,
urlcolor=blue,
]{hyperref}

\def\BibTeX{{\rm B\kern-.05em{\sc i\kern-.025em b}\kern-.08em
    T\kern-.1667em\lower.7ex\hbox{E}\kern-.125emX}}
\begin{document}

\newcommand{\diff}[1]{\noindent\textcolor{black}{#1}}

\title{Benchmarking Mutual Information-based Loss Functions in Federated Learning}

\author{
    \IEEEauthorblockN{
        Sarang S\textsuperscript{*}, 
        Harsh D. Chothani\textsuperscript{*}, 
        Qilei Li\textsuperscript{†}, 
        Ahmed M. Abdelmoniem\textsuperscript{†}, 
        Arnab K. Paul\textsuperscript{*}
    }
    \IEEEauthorblockA{
        \textsuperscript{*}\textit{BITS Pilani, K K Birla Goa Campus, India}
        \textsuperscript{†}\textit{Queen Mary University of London, United Kingdom} \\
        \{f20210966, f20230804, arnabp\}@goa.bits-pilani.ac.in, \{q.li, ahmed.sayed\}@qmul.ac.uk
    }
     \thanks{This work was supported in part by UKRI EPSRC grant EP/X035085/1, BITS Pilani CDRF grant C1/23/173, BioCyTiH grant BBF/BITS(G)/FY2022-23/BCPS-123/24-25/R1, and SERB SRG grant SRG/2023/002445.}
}

\maketitle

\begin{abstract}
    Federated Learning (FL) has attracted considerable interest
    due to growing privacy concerns and regulations
    like the General Data Protection Regulation (GDPR),
    which stresses the importance of privacy-preserving
    and fair machine learning approaches.
    In FL, model training takes place on decentralized data,
    so as to allow clients to upload a locally trained model
    and receive a globally aggregated model
    without exposing sensitive information.
    However,
    challenges related to fairness-such as biases,
    uneven performance among clients,
    and the ``free rider” issue—complicates its adoption.
    In this paper,
    we examine the use of Mutual Information (MI)-based loss functions
    to address these concerns.
    MI has proven to be a powerful method
    for measuring dependencies between variables and optimizing deep learning models.
    By leveraging MI to extract essential features and minimize biases,
    we aim to improve both the fairness and effectiveness of FL systems.
    Through extensive benchmarking,
    we assess the impact of MI-based losses in reducing disparities
    among clients while enhancing the overall performance of FL.
\end{abstract}


\section{Introduction}

Federated Learning (FL) is a paradigm in distributed machine learning (ML) that enables models to be trained cooperatively on decentralized data \cite{mcmahan2017communication}. The relevance of FL has grown significantly in light of stringent privacy regulations like the General Data Protection Regulation (GDPR) \cite{GDPR2016a}. These laws impose strict penalties for violations, underscoring the need for privacy-preserving and fair machine learning solutions. In FL, selected clients receive a global model from a central server called the aggregator, and the clients then train the model locally using their own datasets. The clients send model updates, rather than sensitive data, back to the server, which aggregates them and iteratively enhances the global model~\cite{FL_selection_2023}.

Fairness in federated learning is crucial due to the \textit{black box} nature of machine learning systems, which can result in varied client performance, behaviour, or treatment. Unfairness may arise, for example, if models harbour biases against specific data patterns\cite{biasmitigation23,chang2023biaspropagationfederatedlearning}, if the aggregator doesn't account for differences in clients' hardware or network capacity, or if a client contributes little to training yet benefits from the shared global models (the 'free rider problem')\cite{fraboni21a,zhu2021advanced}. Ensuring fairness directly aligns with GDPR's emphasis on equitable treatment and transparency and balancing trade-offs between performance and fairness remains a persistent challenge in FL systems.
Mutual information (MI) has emerged as a powerful concept for designing loss functions in deep learning, providing a principled way to measure the dependency between variables. MI quantifies the dependency between two random variables \(X\) and \(Y\) by measuring how different the joint distribution of the pair $( X, Y )$ is from the product of the marginal distributions of $X$ and $Y$. MI-based losses enable models to quantify and optimize the dependence between input data and learned representations, capturing salient features while minimizing irrelevant information. They have been successfully employed in tasks such as representation learning, clustering, and domain adaptation, often by maximizing MI between input features and learned representations or minimizing MI between independent features to enforce disentanglement. Efficient estimation of MI in high-dimensional settings, a historically challenging problem, has been addressed by leveraging neural network-based approximations such as MINE (Mutual Information Neural Estimation) \cite{belghazi2018mine} and variational bounds \cite{oord2018representation,poole2019variational}. These advancements enable the practical use of MI as a core objective in model optimization and potentially improve both fairness and performance \cite{hjelm2019learning,choi2022combating}.

In this paper, we explore and benchmark the effectiveness of MI-based loss functions in federated learning,
with a focus on both performance and fairness.
Through comprehensive benchmarking, we assess their potential to reduce disparities
in client outcomes while improving the overall efficiency of FL systems.
Our key contributions include:
(1) Evaluation of state-of-the-art Mutual Information-based loss functions for federated learning.
(2) Analysis and benchmarking of the trade-offs between performance and fairness across different MI-based loss functions.
(3) Empirical results demonstrating that MI-based losses enhance both performance and fairness under diverse FL settings.

\section{Background and Related Work}

In this part, we cover the influence of data distributions on FL training and then introduce Mutual Information as an emerging approach to overcome data heterogeneity. 

\subsection{IID and Non-IID Distributions}

In machine learning, data can be categorized based on distributional assumptions. Independent and Identically Distributed (IID) data assumes that all samples are drawn independently from the same underlying probability distribution. This assumption simplifies analysis, enabling reliable theoretical guarantees like convergence rates and performance bounds \cite{bishop2006pattern,goodfellow2016deep}.
In contrast, Non-Independent and Identically Distributed (Non-IID) data violates either the independence, identical distribution, or both conditions. Non-IID data arises in scenarios where samples exhibit dependencies (e.g., temporal or spatial correlations) or where distributions differ across subsets of data, such as demographic groups or devices \cite{shen2016modeling,kairouz2021advances}.
Challenges in federated learning are amplified with Non-IID data, as clients often hold data with varying distributions. This heterogeneity leads to slower model convergence, performance degradation, and fairness concerns, particularly when certain groups dominate the learning process while others remain underrepresented~\cite{Het_benchmark_2023}. Addressing these challenges requires methods that account for client-level variability and ensure balanced contributions across all distributions \cite{li2020federated,REFL_2023,FLOAT_2024}.

\subsection{Mutual Information}

Mutual information (MI)\diff{\cite{shannon}} measures the dependence between two random variables. It quantifies the reduction in uncertainty about one variable, given knowledge of the other. Formally, MI between two random variables \(X\) and \(Z\) is defined as:
\[
    I(X; Z) = H(X) - H(X \mid Z),
\]
where \(H(X)\) is the Shannon entropy of \(X\), and \(H(X \mid Z)\) is the conditional entropy of \(X\) given \(Z\) . Intuitively, \(I(X; Z)\) measures how much knowing \(Z\) reduces the uncertainty in \(X\)\diff{\cite{CoverThomas2005}}.
Equivalently, MI can also be expressed as the Kullback-Leibler (KL) divergence between the joint distribution \(P_{X,Z}\) and the product of marginals \(P_X \otimes P_Z\):
\[
    I(X; Z) = D_{\text{KL}}(P_{X,Z} \parallel P_X \otimes P_Z).
\]
By modelling the relationships between data distributions across clients, MI can help capture the shared information between client data and the global model.
Two widely used representations of MI based on KL formulation are the Donsker-Varadhan (DV) representation\cite{donsker1975asymptotic} and the Nguyen-Wainwright-Jordan (NWJ) representation\cite{nguyen2010estimating}.
The DV representation is defined as:
\[
    D_{DV}(X, Y) = \sup_{T: \Omega \to \mathbb{R}} \mathbb{E}_{P}[T] - \log\left(\mathbb{E}_{Q}[e^T]\right),
\]
where \( P \) and \( Q \) are probability distributions over the domain \( \Omega \subset \mathbb{R}^d \), and \( T \) is any function. For MI, \( P \) corresponds to the joint distribution \( P_{XY} \), and \( Q \) is the product of the marginals \( P_X P_Y \). The optimal \( T^* \) for this formulation is given by \( T^* = \log\frac{dP}{dQ} + C \), where \( C \) is a constant.
In contrast, the NWJ representation is derived using Fenchel’s inequality\cite{hiriart1996convex}. It is expressed as:
\[
    D_{NWJ}(X, Y) = \sup_{T: \Omega \to \mathbb{R}} \mathbb{E}_{P}[T] - \mathbb{E}_{Q}\left[e^{T-1}\right].
\]
The optimal \( T^* \) in this case is \( \log\frac{dP}{dQ} + 1 \), which differs from the DV formulation due to its self-normalization\cite{belghazi2018mine}.
While both representations are foundational to variational MI estimation, the DV representation guarantees tighter lower bounds for MI, as shown by Ruderman et al. \cite{ruderman2012tightervariationalrepresentationsfdivergences} and Polyanskiy and Wu (2014)\cite{polyanskiy2014information}. Additionally, it has a self-normalizing property, as noted by Belghazi et al. \cite{belghazi2018mine}. These dual representations provide robust theoretical tools for constructing variational bounds on MI.

\vspace{0.5em}
\noindent
\textbf{Variational MI Estimation}
Neural networks have led to the development of various neural network-based variational bounds for mutual information (MI). These bounds are widely used in applications like contrastive learning \cite{oord2018representation,pmlr-v119-chen20j,RepSim_FL_2024} and generative adversarial training \cite{belghazi2018mine,nowozin2016fgantraininggenerativeneural}. Typically, they aim to estimate \( T^* \) using a neural network \( T_\theta: \Omega \to \mathbb{R} \), referred to as the statistics network \cite{belghazi2018mine}, which produces a single real-valued output for input sample pairs.
Leveraging the dual representation of KL divergence, MI can be efficiently estimated and optimized, enabling its integration into loss functions. Although MI-based losses have shown computational tractability, they still struggle with their instability during optimization and training caused by drifting and exploding neural network outputs\cite{choi2022combating}. A simple yet effective regularization has been proposed for the dual representation of KL Divergence to combat the instability\cite{choi2022combating}.
This regularized representation is incorporated into variational MI bounds and used in MI losses designed to balance the maximization of MI, with additional terms to ensure stability and prevent overfitting.

\subsection{Fairness evaluation}

Fairness is a critical challenge in Federated Learning (FL) due to its decentralized nature, where clients with diverse data distributions, participation levels, and computational capabilities collaboratively train a global model. Unlike centralized machine learning, FL amplifies fairness concerns such as unequal model performance, biases against under-represented subgroups, and disproportionate rewards for contributions\cite{kairouz2021advances,mohan2022federated}. These challenges are particularly concerning in high-stakes domains like healthcare and finance, where fairness is not only a technical objective but also a societal and ethical requirement\cite{dwork2012fairness,hardt2016equality}. Ensuring fairness in FL systems is vital to fostering trust among participants and promoting its widespread adoption in sensitive applications.
Traditional definitions of fairness, such as ``fairness through awareness" (treating similar inputs similarly) \cite{dwork2012fairness} and ``group fairness" (ensuring statistical parity for sensitive subgroups)\cite{hardt2016equality}, are partially relevant but insufficient for FL’s unique complexities. The decentralized setup of FL introduces additional layers of heterogeneity, requiring novel fairness metrics and frameworks tailored to the interplay between clients and the server \cite{mohri2019agnostic}. Symptom-based fairness approaches, like those proposed by Vucinich et al. \cite{vucinich2020symptom}, identify measurable indicators of unfairness, such as performance variance across clients, and are intuitive and actionable. Mechanism-driven fairness, such as that defined by Rafi et al. \cite{rafi2021mechanism}, focuses on mitigating unfairness through specific interventions like client resource balancing but may overlook system-wide dynamics like free-riding or malicious behaviour~\cite{Poisining_FL_2025}.
To measure fairness, metrics are required. Despite the additional complexity with FL, some of the centralized metrics to detect algorithmic bias are still insightful in the context of individual client’s performance\cite{barocas2017fairness}. These include ``Equalized Odds" under the umbrella of ``group fairness" that measures the difference in the true positive and false positive rates between two groups, one with and one without a binary sensitive attribute \cite{hardt2016equality}. Uniformity over certain metrics can be indicative of fairness, with the most indicative measure being Jain’s Fairness Index (JFI)\cite{jain1984quantitative}.

\vspace{0.5em}
\noindent
\textbf{Jain's Fairness Index (JFI)} is a bounded, non-linear function based on the coefficient of variation, which is the ratio of the standard deviation to the mean of a population\cite{jain1984quantitative}. The value of J(x) ranges from 0 to 1, where a value of 1 indicates that all clients are performing equally, while a value of \(1/|S_{k}|\) represents the worst-case scenario, where the performance results are entirely non-uniform for a given variable $x$:
\[J(x) = \frac{\left( \sum_{i=1}^{|S_k|} x_i \right)^2}{|S_k| \sum_{i=1}^{|S_k|} x_i^2}\]
JFI quantifies fairness based on the coefficient of variation, which can be accuracy, contribution or the trade-off between two different metrics. Together, these tools enable a nuanced understanding of fairness in FL, addressing individual, group, and global equity in a decentralized and heterogeneous learning environment \cite{kairouz2021advances,Energy_FL_2024}.

\section{Methodology}

This section outlines the experimental setup, datasets, federated learning strategies, loss functions, and other configurations used in this study. The experiments were conducted using simulations implemented in the Flower\cite{beutel2020flower} framework, a flexible and scalable platform for federated learning research.

\subsection{Mutual Information-based losses}

We test three realizations for each representation of Mutual Information based on KL Divergence, as shown in Table \ref{tab:losses}.

We evaluate the performance of mutual information (MI)-based losses using the regularized versions of these MI estimations introduced in Choi et al. \cite{choi2022combating}, including ReMINE (\( I_{\text{MINE}} \)), ReInfoNCE (\( I_{\text{InfoNCE}} \)), ReSMILE (\( I_{\text{SMILE}} \)), ReNWJ (\( I_{\text{NWJ}} \)), ReTUBA (\( I_{\text{TUBA}} \)), ReJS (\( I_{\text{JS}} \)), and ReNWJJS (\( I_{\text{NWJJS}} \)), and compare them against the Cross-Entropy loss as our baseline. We fine-tune the parameters \( \alpha \) and \( \beta \), which control the regularization strength in these losses and select the best-performing values to incorporate into the experimental setup.

\begin{table}[h!]
    \centering
    \caption{MI Representations and Losses. }
    \begin{tabular}{lll}
        \toprule
        \textbf{Representation} & \textbf{Realization} & \textbf{Reference}                                              \\
        \midrule
        \multirow{3}{*}{Donsker-Varadhan}
                                & \(I_\text{MINE}\)     & Belghazi et al. \cite{belghazi2018mine}                          \\
                                & \(I_\text{SMILE}\)    & Song et al. \cite{song2020generativemodelingestimatinggradients} \\
                                & \(I_\text{InfoNCE}\)  & Oord et al. \cite{oord2018representation}                        \\
        \midrule
        \multirow{3}{*}{Nguyen-Wainwright-Jordan}
                                & \(I_\text{NWJ}\)      & Nguyen et al. \cite{nguyen2010estimating}                        \\
                                & \(I_\text{TUBA}\)     & Poole et al. \cite{poole2019variational}                         \\
                                & \(I_\text{JS}\)       & Hjelm et al. \cite{hjelm2019learning}                            \\
        \bottomrule
    \end{tabular}
    \label{tab:losses}
\end{table}

\subsection{Fairness Metrics}
\label{sec:fairness-metrics}

We benchmark the MI-based losses on the federated fairness metrics introduced in Dilley et al. \cite{dilley2024federatedfairnessanalyticsquantifying} to present a complete, symptom-driven (in order to be measurable and independent of the system’s design) and logical definition of fairness.
The following notions define the metrics used; each has an associated \(f\) value, valid \(f \in (0, 1]\) with a value of 1 attributed to complete fairness.

\vspace{0.5em}
\noindent
\textbf{Individual Fairness:}
The measure of client performance is proportionate to their contribution. Using $G$, which is 'Performance Gains' or the ratio of the accuracy of each client after training to their contribution to the global model (measured in Shapely Values)\cite{Shapley1951}, as the coefficient of variability in JFI\cite{jain1984quantitative}, it measures the uniformity of fairness to each individual client.
\[
    f_{j,k} = J(\mathcal{G}),
\]
where \(\mathcal{G} = \{x_{n,k} / s_n\}\), with \(x_{n,k}\) as client \(n\)'s performance in \(k\) round and \(s_n\) their contribution.

\vspace{0.5em}
\noindent
\textbf{Group Fairness:}
Assesses equity across sensitive subgroups defined by attributes such as demographics. Based on Equalized Odds [\(E_n(a)\)] (parity in true and false positive rates between groups) for each client, it evaluates the median of \(E_n(a)\) across all clients to ensure resiliency to outliers.
\[
    f_{g,k} = \text{median}\left( \left\{ \frac{1}{|A|} \sum_{a \in A} E_n(a) \right\} \right),
\]
where \(A\) represents the set of all sensitive attributes, \(E_n(a)\) is the fairness score for sensitive attribute \(a\) and \(f_{g,k}\) aggregates fairness scores across all attributes.

\vspace{0.5em}
\noindent
\textbf{Incentive Fairness:}
Evaluates whether rewards, such as model improvements, are distributed in proportion to contributions. Similar to Individual Fairness, it uses JFI\cite{jain1984quantitative} with $R$, which is the ratio of the reward each client receives, which, in the general case, we take as the accuracy achieved by the global model sent to each local client, to the contribution of each individual client (Shapely Values)\cite{Shapley1951}.
\[
    f_{r,k} = J(R),
\]
where \(R = \{r_{n,k} / s_n\}\).

\vspace{0.5em}
\noindent
\textbf{Orchestrator Fairness:}
This metric works as a measure of the progress of the server's objective, which, in the general federated learning case, we take as the accuracy of the global model by taking the mean of the normalized accuracy across all clients.
\[
    f_{o,k} = \frac{1}{|S_k|} \sum_{n \in S_k} \hat{x}_{n,k},
\]
where \(\hat{x}_{n,k}\) is the normalized accuracy of client \(n\) in round \(k\) ensuring the global model benefits all equitably.
These four notions [\(f_{j}, f_{g}, f_{r}, f_{o}\)] together encompass a complete definition of fairness in federated learning systems\cite{dilley2024federatedfairnessanalyticsquantifying}.

\subsection{Experimental Setup}

We evaluate the performance of mutual information (MI)-based loss functions in federated learning using three federated strategies: FedAvg \cite{mcmahan2017communication}, Ditto \cite{li2021dittofairrobustfederated}, and q-FedAvg \cite{Li2020Fair}. The experiments were performed on the CIFAR-10 dataset, a widely used image classification benchmark, under various scenarios:
(1) Cross-Silo (10 clients) and Cross-Device (100 clients), with a client participation rate of 5\%.
(2) Data partitioning based on IID and non-IID distributions.
The experiments are run over 30 communication rounds, with each client performing 10 local epochs per round, \diff{consistent with Dilley et al. \cite{dilley2024federatedfairnessanalyticsquantifying}}. Each client uses a local train-test split of 90-10, and the model is optimized using the $Adam$ optimizer, ensuring stability during training. 
\diff{All experiments were conducted on a computing node equipped with two NVIDIA RTX 4090 GPUs to ensure efficient training.}
We use the model from the Flower Tutorial\cite{beutel2020flower}. It consists of a simple convolutional neural network with two convolutional layers, one max-pooling layer, and three fully connected layers, as used in FL literature \cite{dilley2024federatedfairnessanalyticsquantifying,mcmahan2017communication}. 

\section{Analysis}

\begin{figure}[!t]
    \centering
    \begin{subfigure}[b]{0.4\textwidth}
        \includegraphics[width=\textwidth]{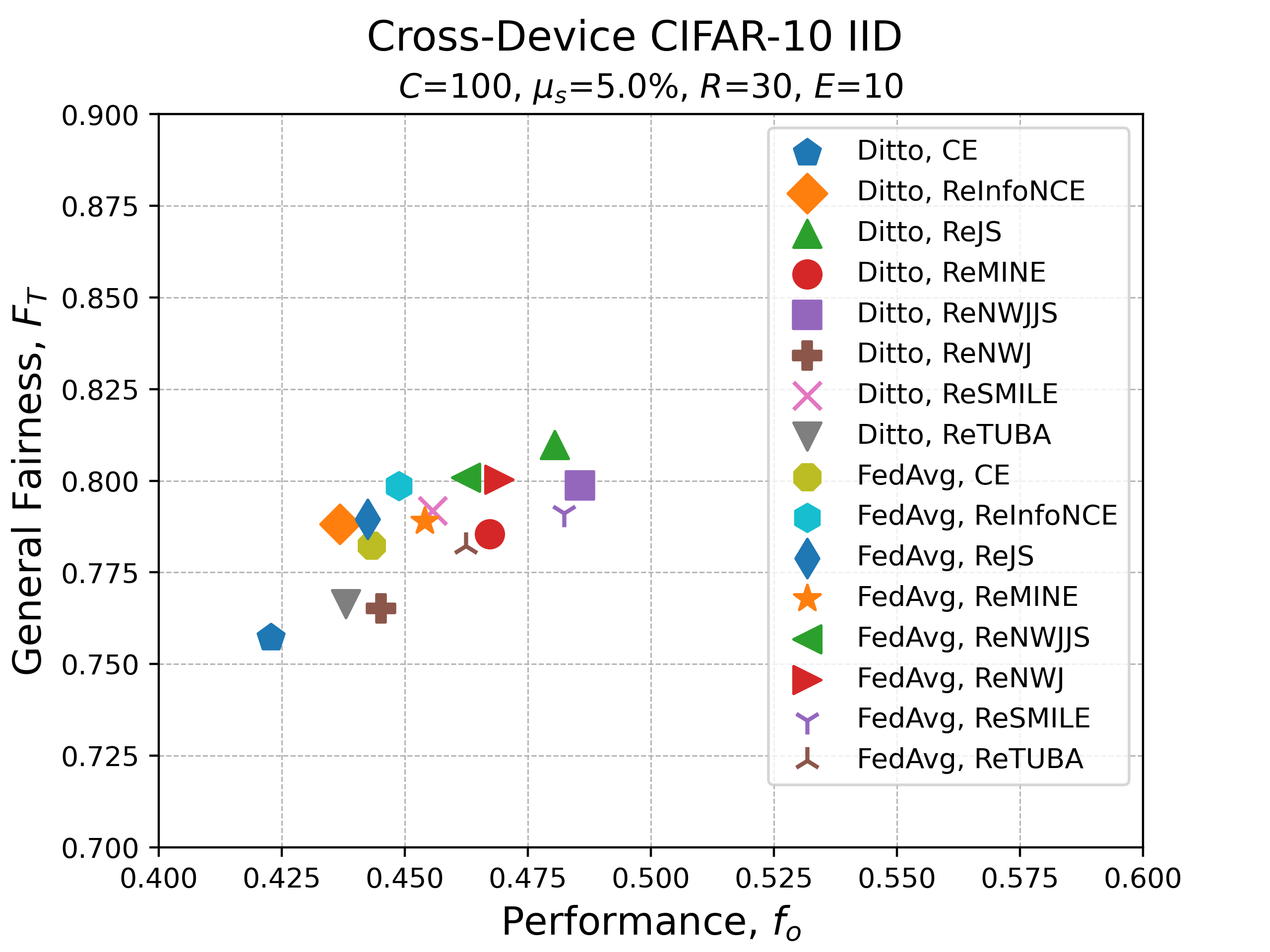}
        \caption{IID across FedAvg and Ditto}
        \label{fig:cross-device-iid}
    \end{subfigure}
    \hfill
    \begin{subfigure}[b]{0.4\textwidth}
        \includegraphics[width=\textwidth]{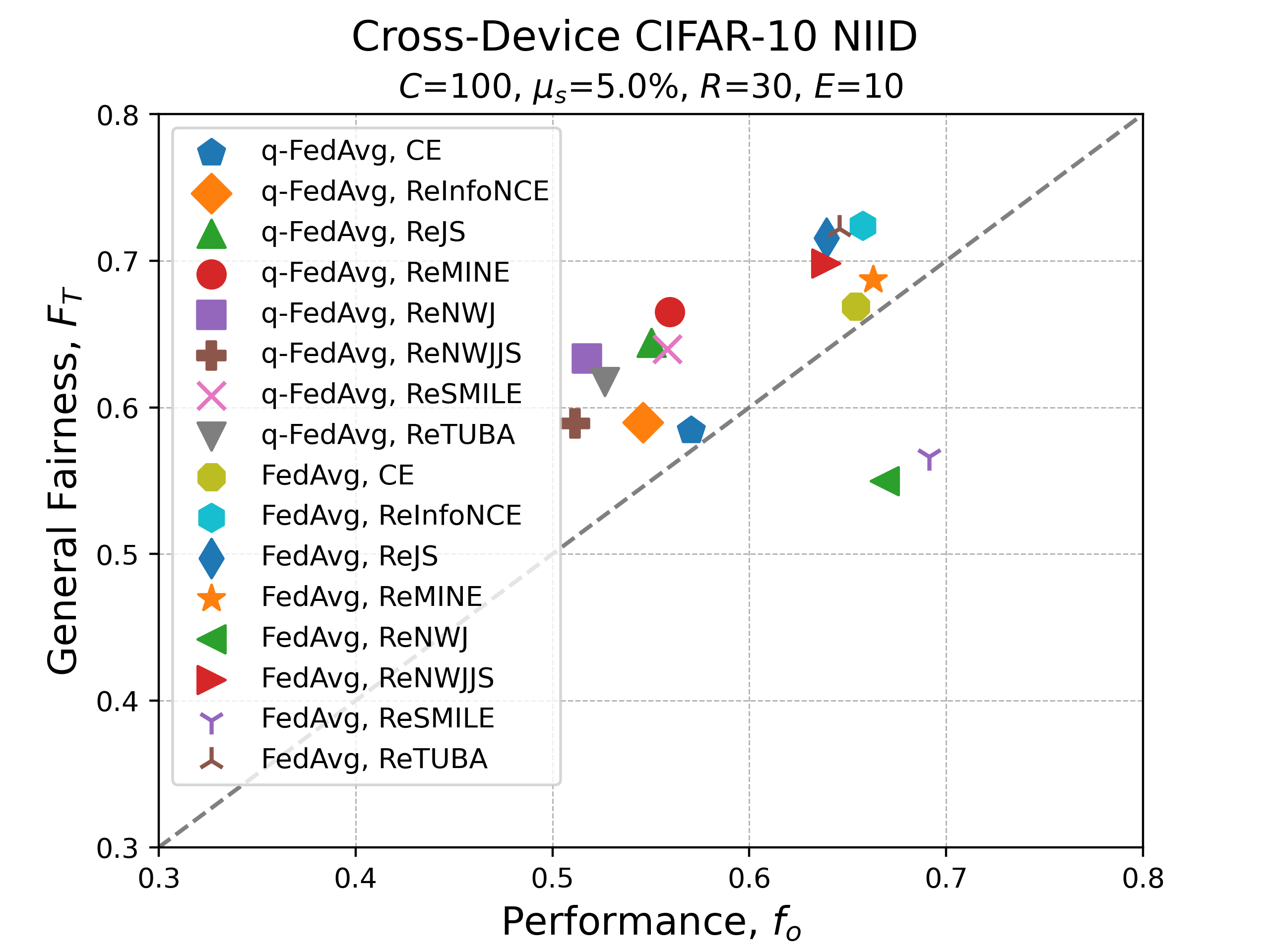}
        \caption{Non-IID across FedAvg and qFedAvg}
        \label{fig:cross-device-niid}
    \end{subfigure}
    \caption{Cross-Device FL in IID \& Non-IID settings.}
    \label{fig:combined-cross-device}
\end{figure}

We present a comprehensive evaluation of various MI-based loss functions, examining their impact on both performance and fairness across different federated learning algorithms and data distribution scenarios. Our analysis focuses on two key metrics: General Fairness and Performance (Orchestrator Fairness).
General Fairness ($F_t$) measures system fairness by combining multiple fairness interpretations described in Section \ref{sec:fairness-metrics}. It is computed as the arithmetic mean of four distinct fairness components:
\[
    F_t = (f_j + f_g + f_r + f_o)/4
\]
where $f_j$, $f_g$, $f_r$, and $f_o$ represent Individual fairness, Group fairness, Incentive fairness, and Orchestrator fairness, respectively.
Performance, measured through Orchestrator Fairness, captures the system's overall effectiveness from the aggregator's perspective by computing the average performance across all clients. To understand the practical implications of different loss functions, we analyze the inherent trade-off between General Fairness and Performance across various experimental settings. This approach allows us to identify solutions that achieve high fairness without significantly compromising system performance.

\subsection{Does data distribution have an effect on fairness?}
Our experimental results reveal a significant impact of data distribution on fairness outcomes in federated learning. We observe consistently higher general fairness values in IID scenarios compared to Non-IID scenarios across Cross-Device (Fig. \ref{fig:combined-cross-device}) and Cross-Silo (Fig. \ref{fig:combined-cross-silo}) settings. This pattern suggests that achieving fairness is more straightforward when data is independently and identically distributed across clients. This is likely because IID inherently represents a fair data distribution, regardless of the chosen loss function.

\noindent
\textbf{Performance of MI-based losses in IID settings.}
In IID environments, MI-based loss functions demonstrate remarkable performance and fairness across FedAvg and Ditto algorithms. Figure \ref{fig:cross-device-iid} shows that in cross-device scenarios, losses such as $I_\text{ReJS}$ and $I_\text{ReMINE}$ extract both higher performance and fairness from the Ditto algorithm. In both cross-device and cross-silo settings, $I_\text{ReNWJ}$ and $\text{ReSMILE}$ achieve higher fairness and performance for FedAvg.

\noindent
\textbf{Performance of MI-based losses in Non-IID settings.}
While Non-IID scenarios present more challenging conditions, MI-based loss functions maintain their effectiveness. In the cross-device Non-IID setting (Fig. \ref{fig:cross-device-niid}), the combination of FedAvg with $I_\text{ReInfoNCE}$ achieves \~0.5 general fairness over $Cross Entropy$ while matching the performance, and similar trends can be observed for $q-FedAvg$  as well. An impressive jump in general fairness can be observed in the case of Ditto ($I_\text{ReInfoNCE}$ (0.725) vs $CE$ (0.5)) in Fig. \ref{fig:cross-silo-niid}. These results demonstrate the adaptability of MI-based approaches to heterogeneous data distributions.

\subsection{Are the benefits consistent across different FL scenarios?}
\begin{figure}[!t]
    \begin{subfigure}[b]{0.4\textwidth}
        \includegraphics[width=\textwidth]{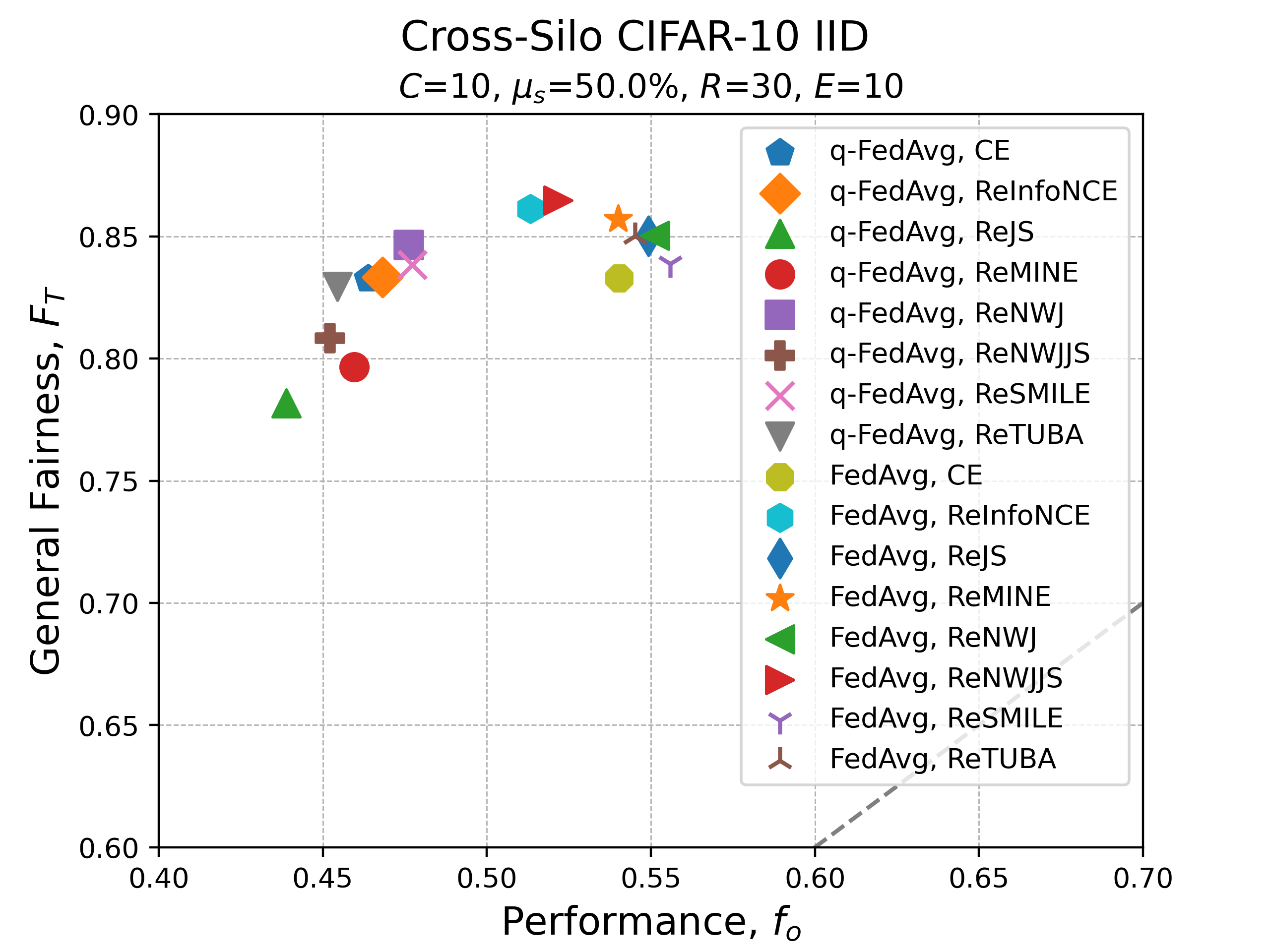}
        \caption{IID across FedAvg and qFedAvg}
        \label{fig:cross-silo-iid}
    \end{subfigure}
    \hfill
    \begin{subfigure}[b]{0.4\textwidth}
        \includegraphics[width=\textwidth]{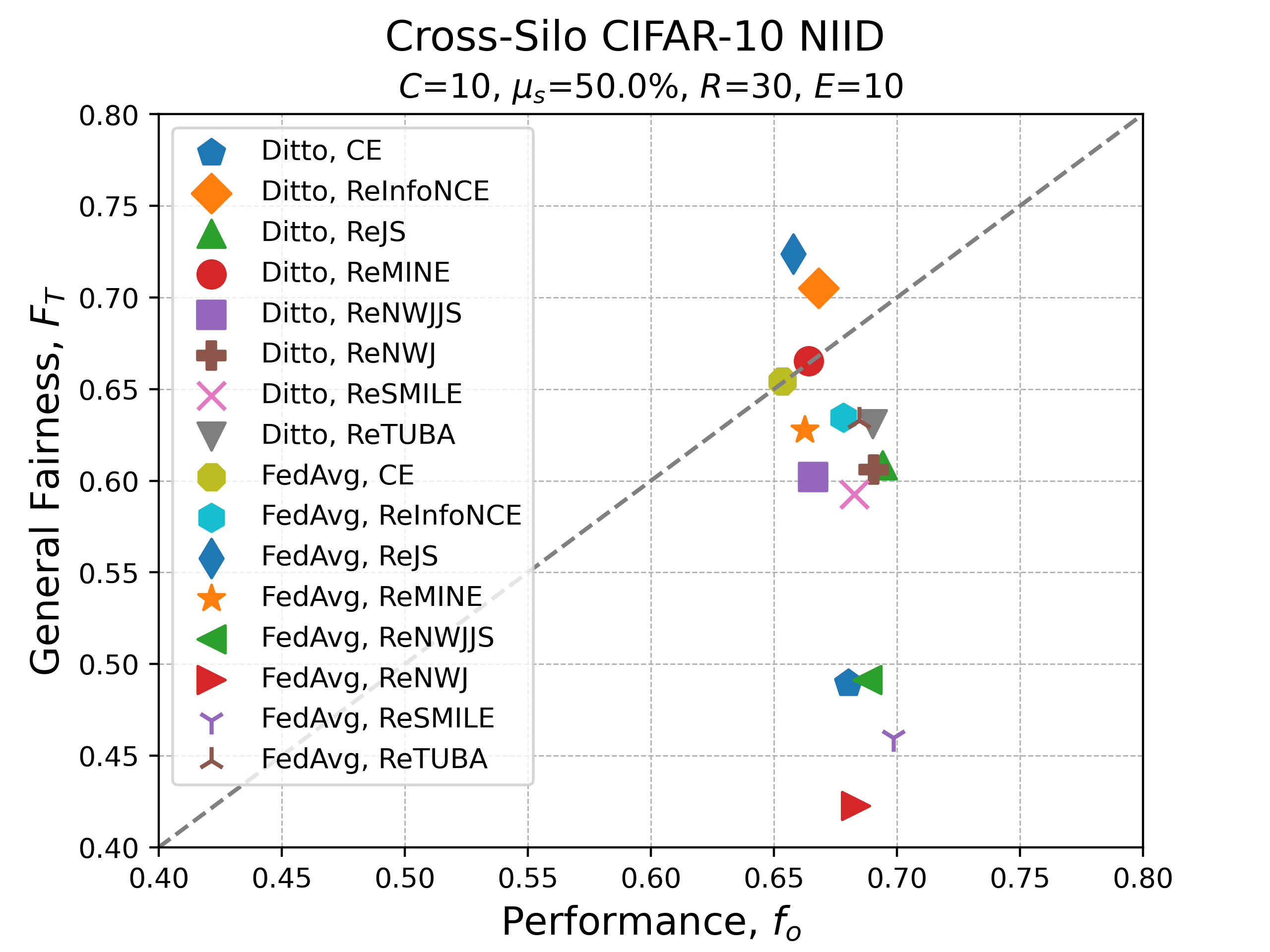}
        \caption{Non-IID across FedAvg and Ditto}
        \label{fig:cross-silo-niid}
    \end{subfigure}
    \caption{Cross-Silo FL in IID and Non-IID settings.}
    \label{fig:combined-cross-silo}
\end{figure}

The fairness-performance trade-offs show remarkable consistency between cross-device and cross-silo settings, as demonstrated by comparing Figures \ref{fig:cross-device-iid} and \ref{fig:cross-silo-iid} (IID) and Figures \ref{fig:cross-device-niid} and \ref{fig:cross-silo-niid} (Non-IID), indicating that MI-based loss functions offer scalable benefits across various federated learning scenarios and client configurations.

\subsection{Breaking down the fairness metrics}

\begin{figure}[h!]
    \centering
    \begin{subfigure}[b]{0.225\textwidth}
        \includegraphics[width=\textwidth]{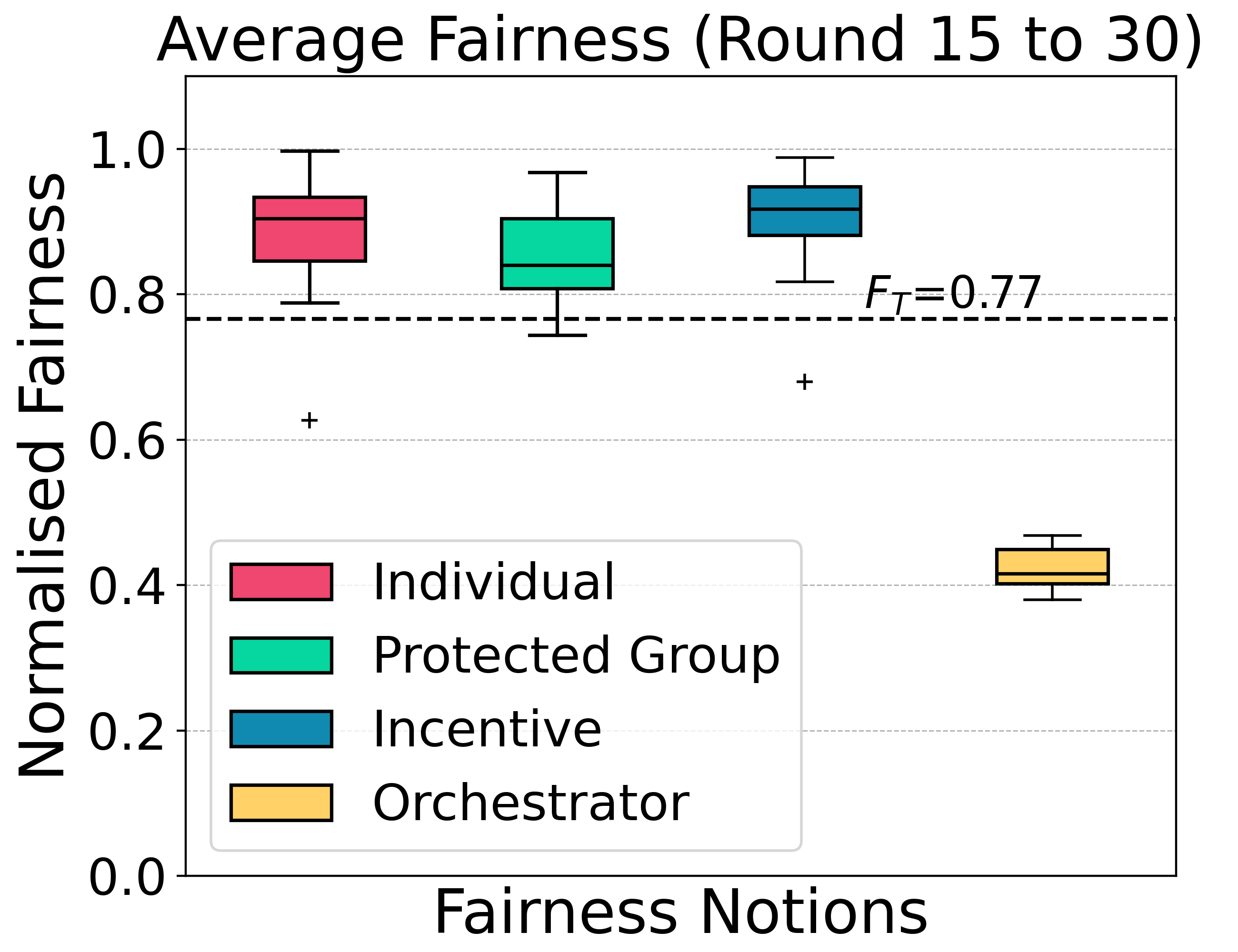}
        \caption{CE average}
        \label{fig:ditto-ce-cross-device-bar-average}
    \end{subfigure}
    \hfill
    \begin{subfigure}[b]{0.225\textwidth}
        \includegraphics[width=\textwidth]{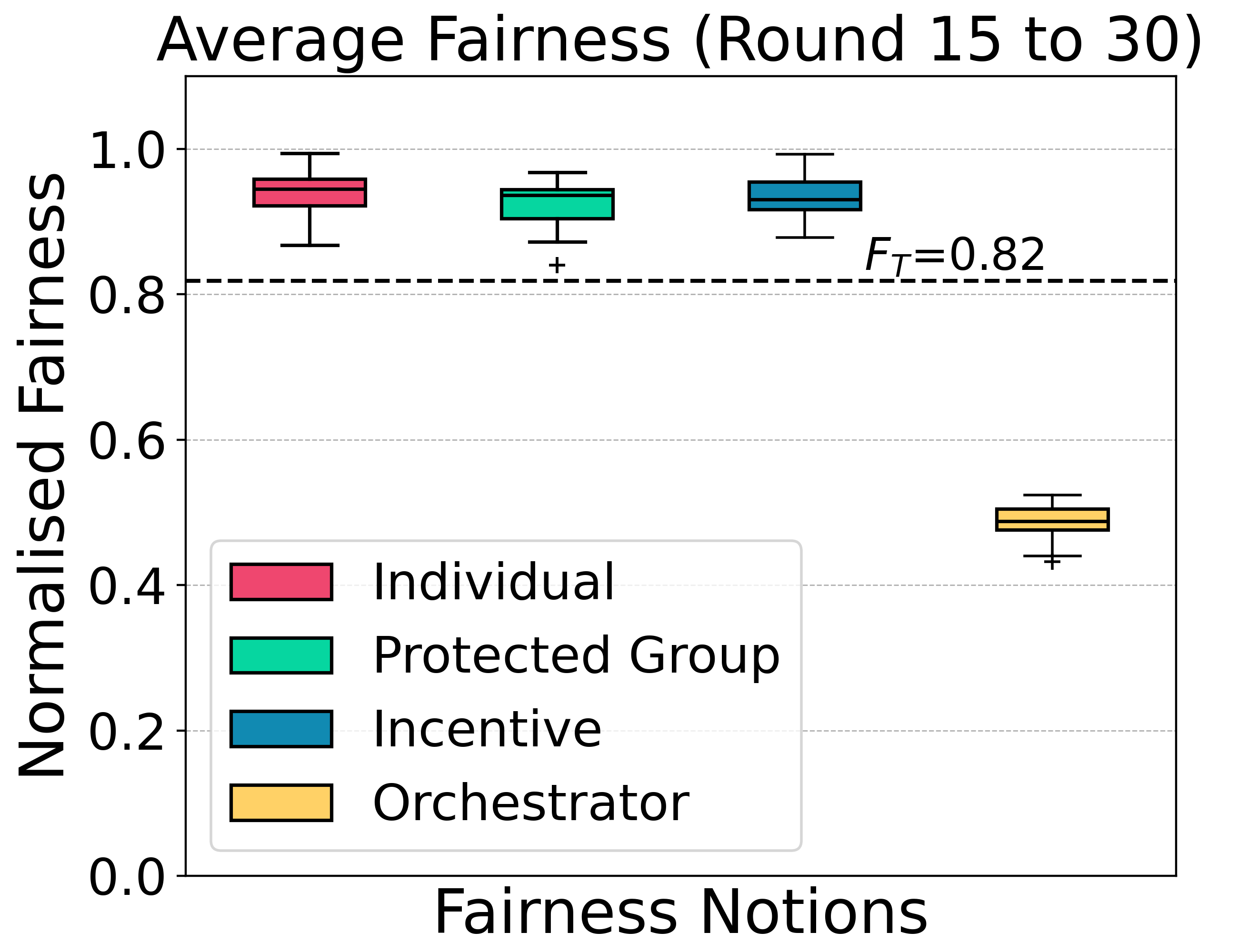}
        \caption{$I_\text{ReJS}$ average}
        \label{fig:ditto-rejs-cross-device-average}
    \end{subfigure}
    \begin{subfigure}[b]{0.225\textwidth}
        \includegraphics[width=\textwidth]{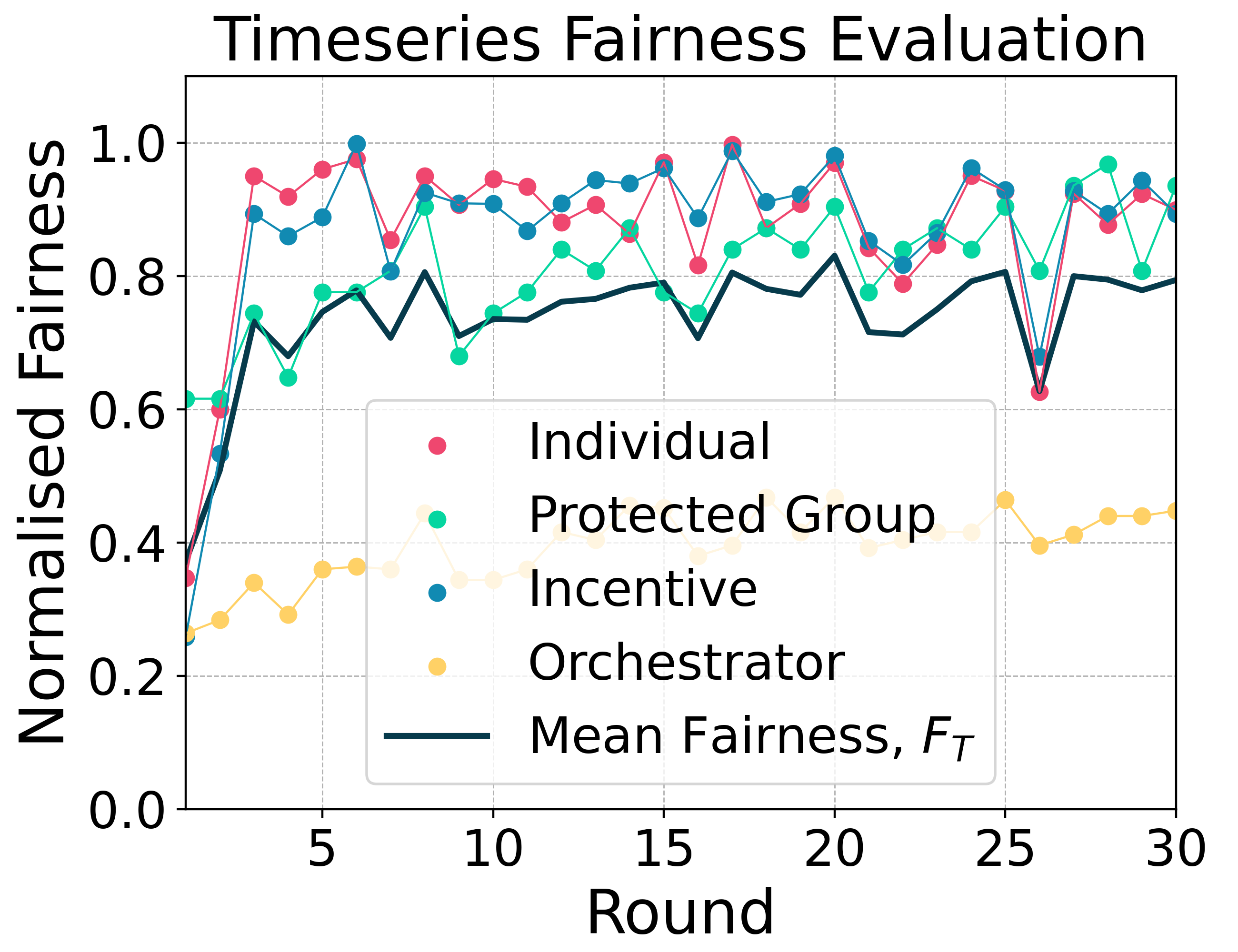}
        \caption{CE timeseries}
        \label{fig:ditto-ce-cross-device-timeseries}
    \end{subfigure}
    \hfill
    \begin{subfigure}[b]{0.225\textwidth}
        \includegraphics[width=\textwidth]{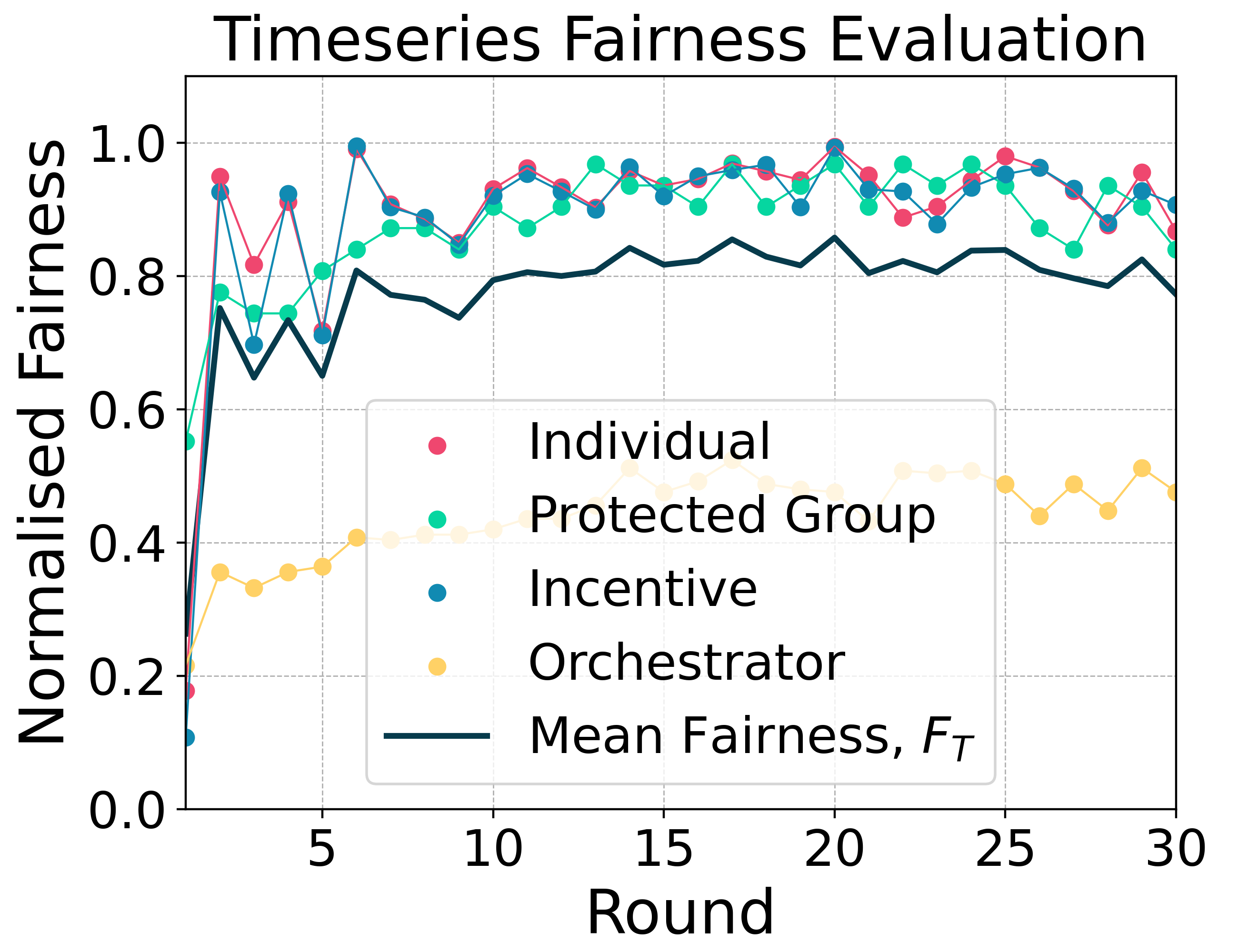}
        \caption{$I_\text{ReJS}$ timeseries}
        \label{fig:ditto-rejs-cross-device-timeseries}
    \end{subfigure}
    \caption{Comparison between Ditto cross-device IID experiments for CE and \protect $I_\text{ReJS}$.}
    \label{fig:combined-iid}
\end{figure}

\begin{figure}[h!]
    \centering
    \begin{subfigure}[b]{0.225\textwidth}
        \includegraphics[width=\textwidth]{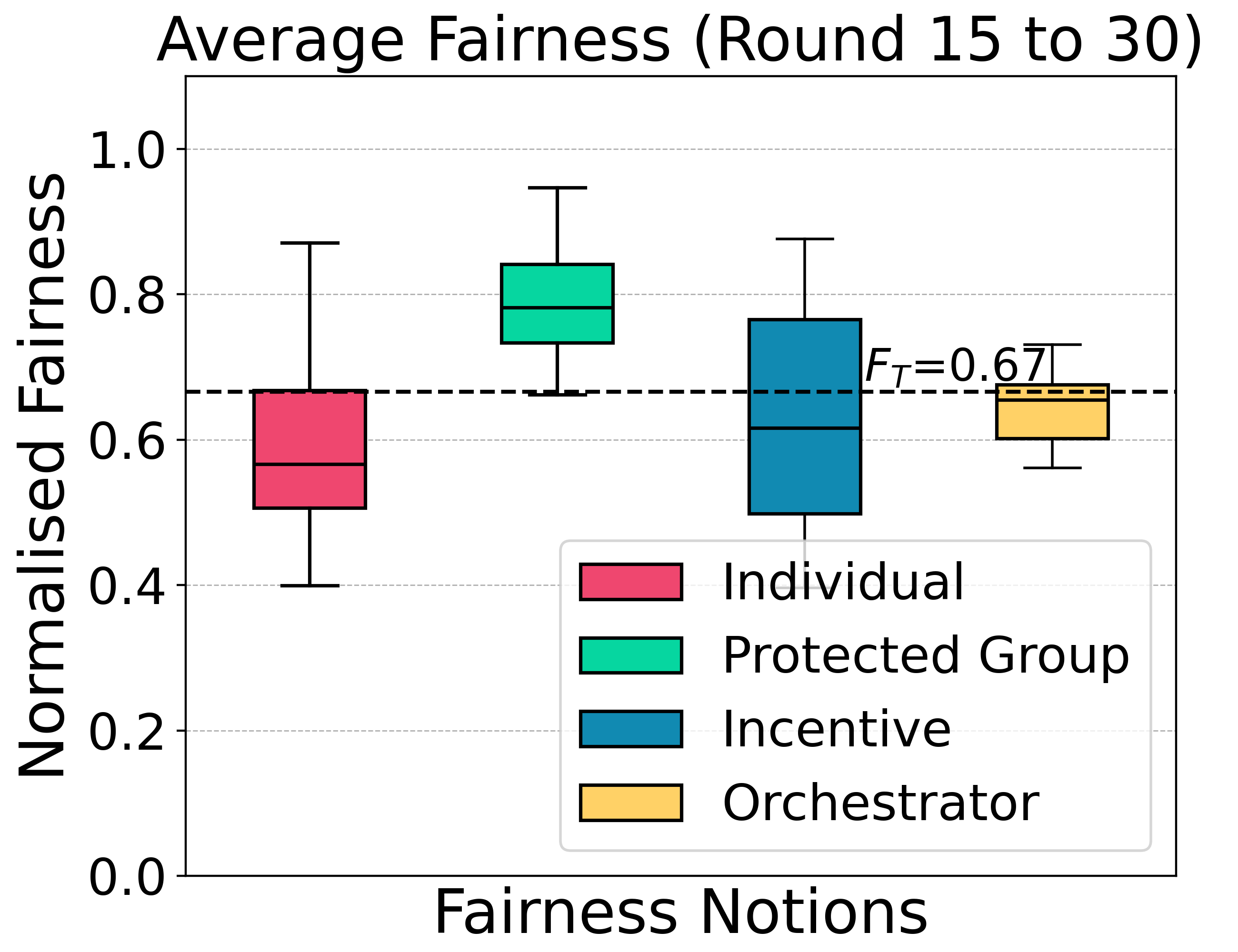}
        \caption{CE average}
        \label{fig:fedavg-ce-cross-device-bar-average}
    \end{subfigure}
    \hfill
    \begin{subfigure}[b]{0.225\textwidth}
        \includegraphics[width=\textwidth]{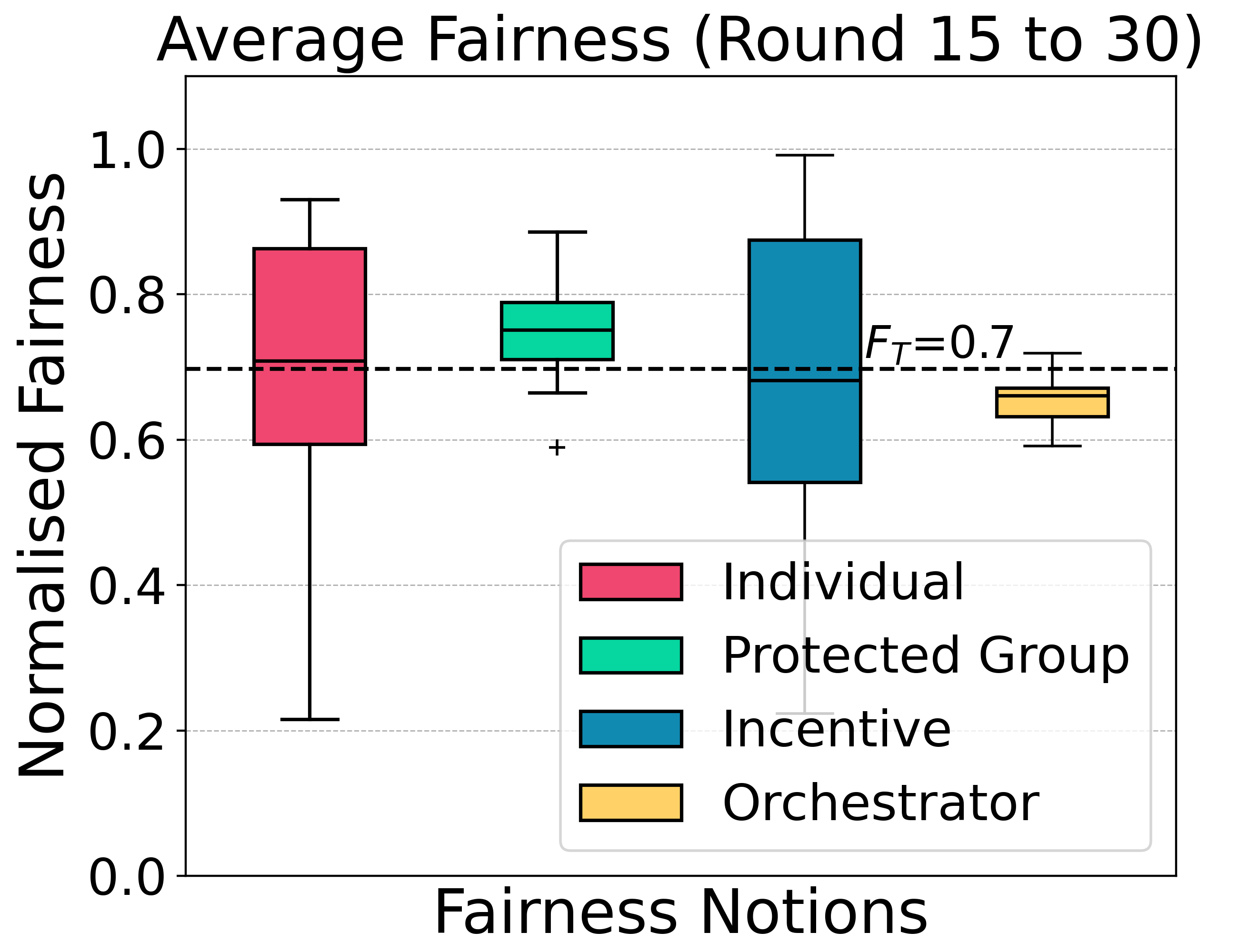}
        \caption{$I_\text{ReInfoNCE}$ average}
        \label{fig:fedavg-reinfonce-cross-device-average}
    \end{subfigure}
    \begin{subfigure}[b]{0.225\textwidth}
        \includegraphics[width=\textwidth]{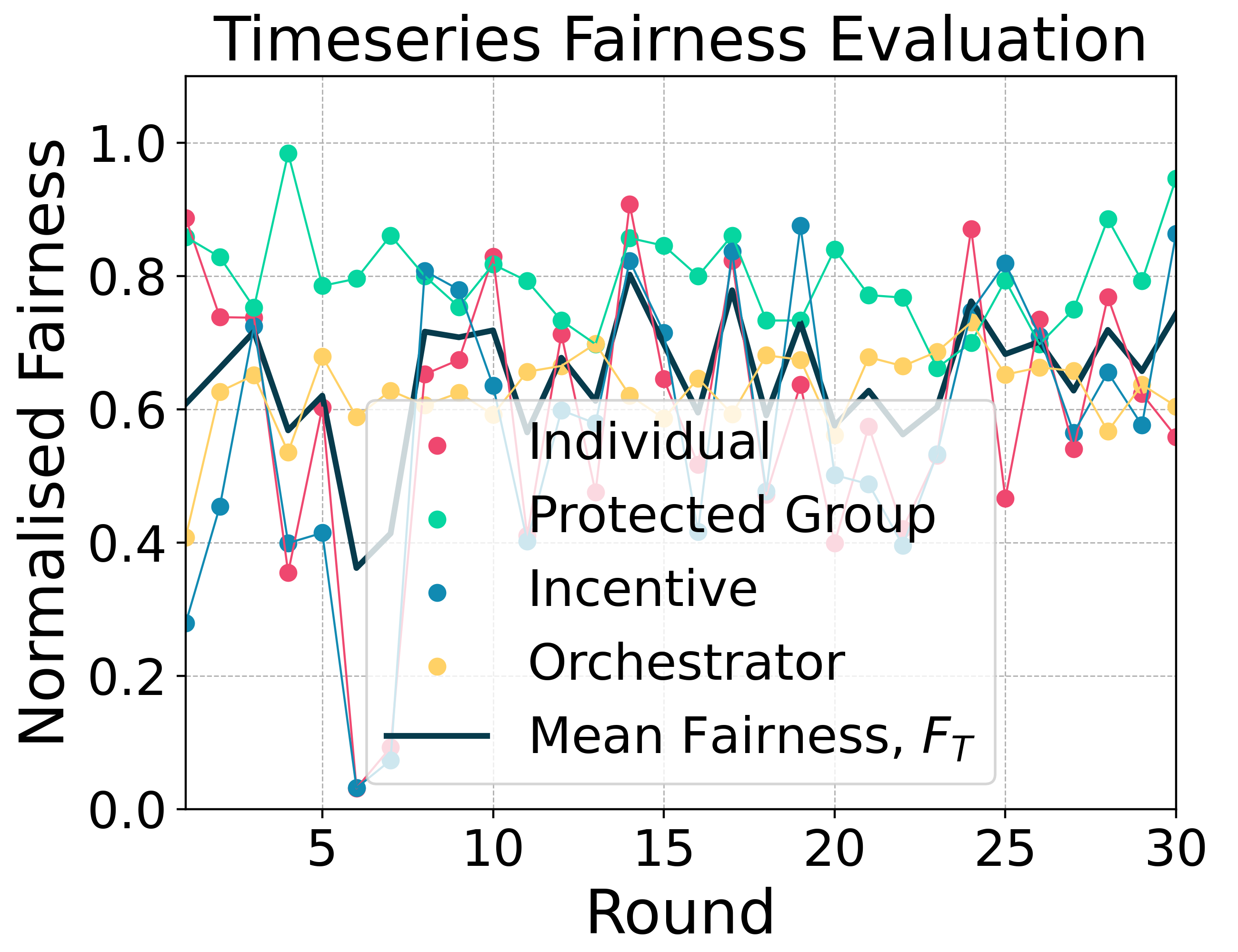}
        \caption{CE timeseries}
        \label{fig:fedavg-ce-cross-device-timeseries}
    \end{subfigure}
    \hfill
    \begin{subfigure}[b]{0.225\textwidth}
        \includegraphics[width=\textwidth]{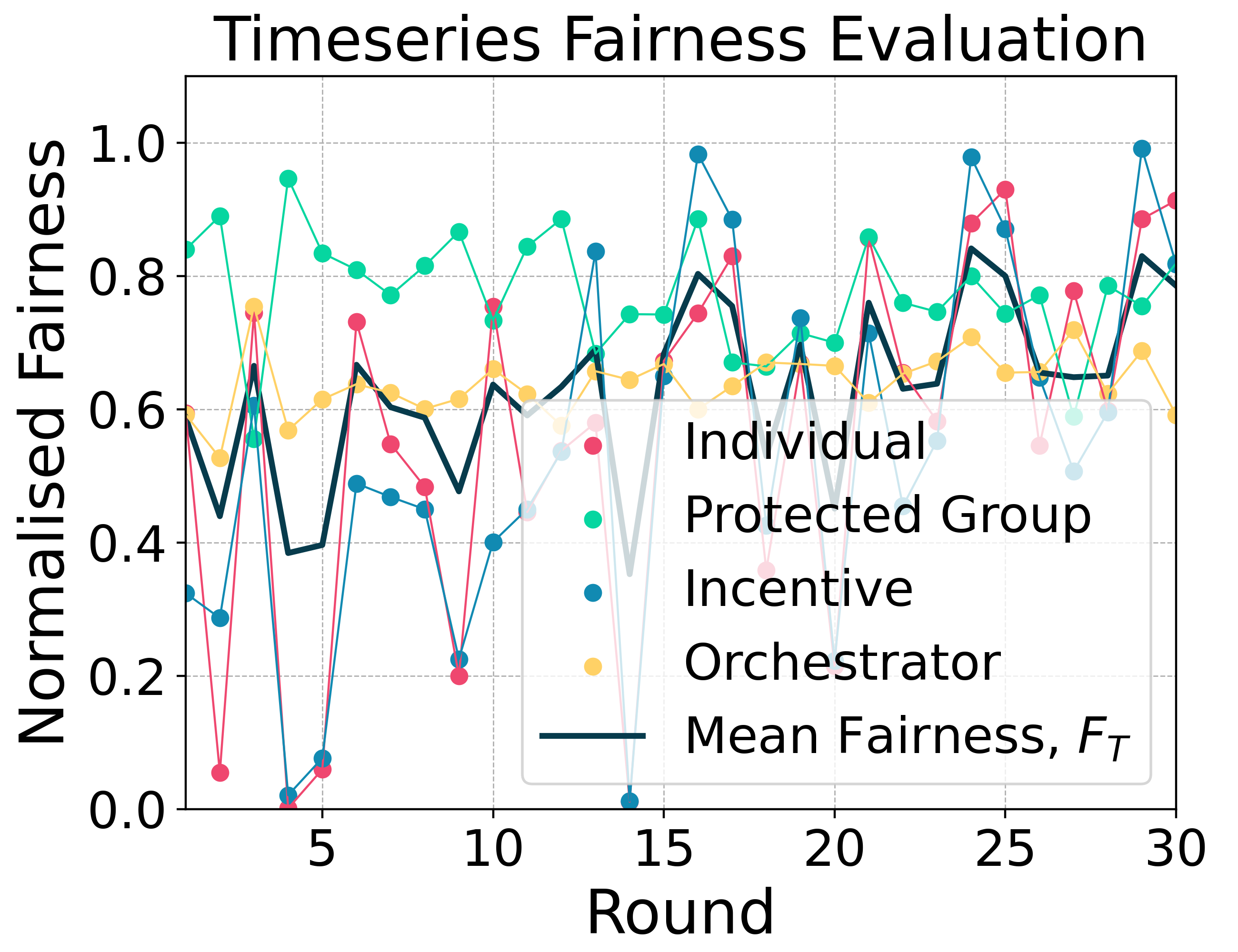}
        \caption{$I_\text{ReInfoNCE}$ timeseries}
        \label{fig:fedavg-reinfonce-cross-device-timeseries}
    \end{subfigure}
    \caption{Comparison between FedAvg cross-device Non-IID experiments for CE and $I_\text{ReInfoNCE}.$}
    \label{fig:combined-niid}
\end{figure}

Figure \ref{fig:combined-iid} shows the component-wise breakdown of the fairness metrics for $CE$ and $I_\text{ReJS}$ for the Ditto algorithm, which clearly illustrates the higher mean and lower variance for each of the four components of $I_\text{ReJS}$, which leads to its overall higher general fairness.
Figure \ref{fig:combined-niid} shows the component-wise breakdown of the fairness metrics for $CE$ and $I_\text{ReInfoNCE}$ using FedAvg in a Non-IID setting. This clearly illustrates higher mean and maximum values for Individual and Incentive Fairness as well as lower variance for Protected Group and Orchestrator Fairness using $I_\text{ReInfoNCE}$. Although this ensures overall higher general fairness, it is worth noting that in this particular setting, $CE$ has a marginally higher mean in Group Fairness and lower variance in Incentive fairness, ensuring more consistency across those metrics.
Our results demonstrate that MI-based loss functions can effectively enhance fairness in federated learning systems without significantly compromising performance. This is particularly noteworthy in challenging Non-IID scenarios, where traditional approaches often struggle to maintain client fairness. The consistent performance across different architectures and data distributions suggests that MI-based loss functions provide a robust solution for fairness-aware federated learning.

\section{Conclusion}
In this paper, we explored the effectiveness of Mutual Information (MI)-based loss functions in federated learning
by focusing on fairness and performance.
Our results show that MI-based losses improve fairness across different FL strategies and data distributions, particularly in Non-IID settings where traditional approaches struggle.
Notably, losses like \( I_{\text{ReJS}} \) and \( I_{\text{ReInfoNCE}} \) achieved higher fairness while maintaining strong performance.
While MI-based losses introduce computational overhead, they offer a promising direction for fairness-aware FL. \diff{Future work can optimize their efficiency and explore hybrid approaches that introduce task-specific priors for higher fairness}. Overall, MI-based losses provide a viable solution for improving fairness in decentralized learning.

\bibliographystyle{unsrt}
\bibliography{refs}

\end{document}